\documentclass{article}
\usepackage{spconf}
\ninept 

\usepackage[utf8]{inputenc} 
\usepackage[T1]{fontenc}    

\usepackage{url}            
\usepackage{booktabs}       
\usepackage{amsfonts}       
\usepackage{nicefrac}       
\usepackage{microtype}      
\usepackage{siunitx}
\usepackage{float}
\usepackage{caption}
\usepackage{soul}
\usepackage{enumitem}

\usepackage{lipsum}

\usepackage{hyperref}       
\hypersetup{allcolors=blue,colorlinks=true}
\usepackage{graphicx}
\usepackage{amsmath}
\usepackage{amsthm}
\usepackage{amssymb}
\usepackage{amsfonts}
\usepackage{mathrsfs}
\usepackage{multirow}
\usepackage{algorithm}
\usepackage{subfigure}
\usepackage{thm-restate}
\usepackage{mathtools}
\usepackage{comment}
\usepackage{multicol}
\usepackage{bm}
\newcommand{\R}{\mathbb{R}}

\newcommand{\bI}{\bm{I}}
\newcommand{\bS}{\bm{S}}

\newcommand{\bC}{\bm{C}}
\newcommand{\bL}{\bm{L}}

\newcommand{\bU}{\bm{U}}
\newcommand{\bV}{\bm{V}}

\newcommand{\bX}{\bm{X}}

\newcommand{\bOmega}{\bm{\Omega}}
\newcommand{\bu}{\bm{u}}
\newcommand{\bv}{\bm{v}}

\newcommand{\V}{\mathcal{V}}

\usepackage{bbm}

\newcommand{\bzero}{\mathbf{0}}

\DeclareMathOperator*{\argmin}{arg\,min}
\DeclareMathOperator*{\argmax}{arg\,max}
\DeclareMathOperator{\prox}{\textsf{prox}}

\DeclareMathOperator{\Tr}{Tr}

\usepackage{pifont}

\usepackage{thm-restate}

\usepackage[natbib=true,
            bibstyle=ieee,
            citestyle=numeric-comp,
            uniquename=false,
            uniquelist=false,
            maxcitenames=2,
            mincitenames=1,
            maxbibnames=6,
            mincrossrefs=99]{biblatex}
\renewbibmacro{in:}{\ifentrytype{article}{}{\printtext{\bibstring{in}\intitlepunct}}}
\DefineBibliographyStrings{english}{url = Available at}

\AtEveryBibitem{%
  \clearlist{publisher}%
  \clearfield{pagetotal}%
  \clearfield{series}%
  \clearfield{edition}%
  \clearfield{isbn}%
  \clearname{editor}%
  \clearlist{location}%
}

\title{Sparse Partial Least Squares for Coarse Noisy Graph Alignment}

\makeatletter
\def\twoauthors#1#2#3#4{\gdef\@address{}
   \gdef\@name{\begin{tabular}{@{}c@{}}
        {\em #1} \\ \\
        #2\relax
   \end{tabular}\hskip 0.5in\begin{tabular}{@{}c@{}}
        {\em #3} \\ \\
        #4\relax
\end{tabular}}}
\makeatother

\twoauthors{Michael Weylandt\sthanks{MW's research is supported by an appointment to the Intelligence Community Postdoctoral Research Fellowship Program at the University of Florida Informatics Institute, administered by Oak Ridge Institute for Science and Education through an interagency agreement between the U.S. Department of Energy and the Office of the Director of National Intelligence.}%
            and George Michailidis\sthanks{GM was supported in part by NSF grants DMS 1830175 and DMS 1821220.}}%
            {University of Florida Informatics Institute \\
             University of Florida \\
             Gainesville, FL 32611 \\
             \href{mailto:michael.weylandt@ufl.edu}{michael.weylandt@ufl.edu}  \href{mailto:gmichail@ufl.edu}{gmichail@ufl.edu}}%
             {T. Mitchell Roddenberry\sthanks{TMR was partially supported by the 2020/21 Exxon-Mobil Graduate Fellowship awarded by the Ken Kennedy Institute at Rice University.}}%
             {Department of Electrical and Computer Engineering \\
              Rice University\\
              Houston, TX 77005 \\
              \href{mailto:mitch@rice.edu}{mitch@rice.edu}}

\begin{document}
\begin{refsection}
\maketitle


\begin{abstract}
Graph signal processing (GSP) provides a powerful framework for analyzing signals arising in a variety of domains. In many applications of GSP, multiple network structures are available, each of which captures different aspects of the same underlying phenomenon. To integrate these different data sources, graph alignment techniques attempt to find the best correspondence between vertices of two graphs. We consider a generalization of this problem, where there is no natural one-to-one mapping between vertices, but where there is correspondence between the community structures of each graph. Because we seek to learn structure at this higher community level, we refer to this problem as ``coarse'' graph alignment.  To this end, we propose a novel regularized partial least squares method which both incorporates the observed graph structures and imposes sparsity in order to reflect the underlying block community structure. We provide efficient algorithms for our method and demonstrate its effectiveness in simulations.

\noindent 
\end{abstract}

\section{Introduction}
Consider signals arising simultaneously on two graphs $\mathcal{G}_1, \mathcal{G}_2$ of size $n_1, n_2$, respectively. These graphs may represent two distinct social networks with graph signals capturing the amount of discussion of a particular topic on a given day. While different users may populate each social network with potentially no overlap, the patterns of discussion are likely to be correlated, particularly within communities of common interest. For example, the communities of politically-engaged Facebook and Twitter users are both likely to actively discuss the same breaking news stories at the same time, even if there is no direct ``cross-talk'' between the two networks. We consider an idealized version of this situation, where $m$ paired signals, sampled independently, are observed on the networks $\mathcal{G}_1$ and $\mathcal{G}_2$. Many existing popular community detection methods (for a single network, or networks with a shared set of nodes) examine low-rank approximations of the covariance matrix of graph signals~\citep{Schaub:2020,Roddenberry:2020}: intuitively, if two nodes are in the same community, their responses to external stimuli, represented as graph signals, are likely to be highly correlated and so methods that capture the major patterns of covariance are likely to identify communities. 

Our approach extends this intuition to the multi-graph setting: rather than investigating the covariance of signals supported on a single graph, we consider a low rank model for the matrix of inner products of each node pair: $\bX_1^{\top}\bX_2$ where $\bX_1 \in \R^{m \times n_1}, \bX_2 \in \R^{m \times n_2}$ are data matrices of the $m$ graph signals observed on graphs $\mathcal{G}_1, \mathcal{G}_2$ respectively. We extend the classical PLS approach to this problem in two directions: first, we assume that the covariance among signals within a graph reflects the community structure within that graph, hence we seek PLS vectors which are smooth with respect to the observed graph structures. Second, we directly incorporate sparsity into the estimated PLS vectors to get 
community assignments.

\subsection{Background: Graph Signal Processing}

A \emph{graph signal} is a function mapping the nodes $\mathcal{V}$ of a graph $\mathcal{G}=(\mathcal{V},\mathcal{E})$ to the real numbers.
By indexing the nodes $\mathcal{V}$ with the integers $\{1,2,\ldots,n\}$, where $n=|\mathcal{V}|$, we represent a graph signal $x:\mathcal{V}\to\mathbb{R}$ as a vector $\mathbf{x}\in\mathbb{R}^n$, where $x_i$ in the vector representation is equal to $x(i)$ in the functional representation.
We say that a graph signal $\mathbf{x}$ is smooth if it has little energy in the quadratic form of the \emph{graph Laplacian} $\mathbf{L}$, \emph{i.e.,} $\mathbf{x}^\top\mathbf{L}\mathbf{x}/\|\mathbf{x}\|_2^2$ has a small value.
%
%
We refer the reader to the excellent survey by \citet{Ortega:2018} for a more complete introduction to the subject.

\subsection{Background: Graph Alignment Problems}

The problem of \emph{graph alignment} takes two graphs $\mathcal{G}_1=(\mathcal{V}_1,\mathcal{E}_1), \mathcal{G}_2=(\mathcal{V}_2,\mathcal{E}_2)$, and seeks a bijection between the node sets $\mathcal{V}_1$ and $\mathcal{V}_2$ that maximally preserves the edge structure. In this work, we are primarily concerned with methods related to \emph{spectral graph alignment} techniques, which leverage the eigenvalue decompositions of graph matrices in order to find a useful approximate alignment~\citep{Umeyama:1988,Fan:2020}. 

\citet{Chen:2018-GraphCCA,Chen:2019-GraphCCA} have considered a graph-regularized approach to canonical correlation analysis (CCA), wherein graph-structured dependencies between observations are used to enforce graph-smoothness in the canonical variables. The proposed approach shares similarities, but also exhibits marked differences; the former method assumes a known and pre-aligned single graph, common to each data set, and uses it to improve prediction of different blocks of correlated features. In contrast, we have multiple graphs - unaligned and indeed potentially unalignable - and seek to identify community structures within each graph and meaningful mappings between those estimated structures. 

We highlight two recent and closely-related lines of work. \citet{Maretic:2020} relax the classical graph alignment problem to allow for ``many-to-many'' mappings and then leverages Wasserstein metrics to further relax the problem to incorporate mappings between graphs of different size; compared to our approach, it continues to work at the level of vertex, not community, correspondences and only considers community detection as a downstream task. \citet{Zhou:2019,Zhou:2020} consider spectral clustering under the ``bipartite'' stochastic block model, which extends the traditional SBM to paired networks; compared to our approach, they do not take advantage of graph signals and restrict their analysis to a specific SBM model.

\begin{figure}[ht]
\centering
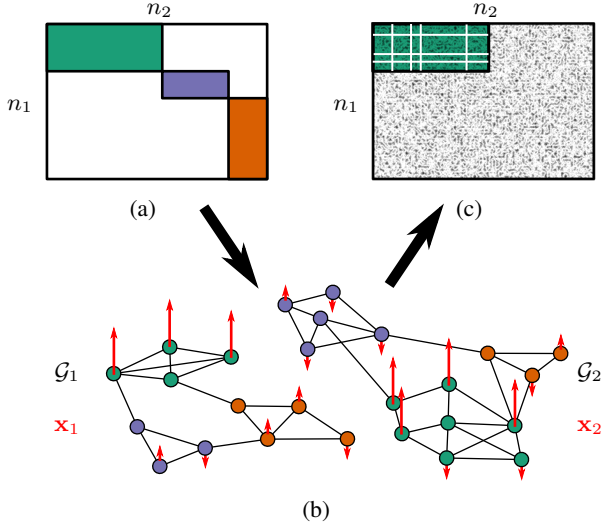
\vspace{-0.4in}
\caption{A schematic of our model: (a) Joint community membership matrix (unobserved) for graphs $\mathcal{G}_1,\mathcal{G}_2$ over $n_1,n_2$ nodes, respectively.
(b) Noisy graph signals $\mathbf{x}_1,\mathbf{x}_2$ (red arrows) are observed on graphs $\mathcal{G}_1,\mathcal{G}_2$, which consist of disjoint node sets, but with correlated communities (green, purple, and orange). (c) Outer products of graph signals $\mathbf{x}_1\mathbf{x}_2^\top$ yield noisy low-rank approximations of the hidden joint community matrix. For sufficiently large sample sizes, low-rank approximations of the sum of these matrices can recover the true joint community membership matrix.}
\label{fig:schematic}
\end{figure}

\subsection{Background: Partial Least Squares}

The proposed approach builds upon the classical method of partial least squares: PLS is a form of rank-reduced multi-response regression that aims to identify a shared low-dimensional representation of two or more data sets. Due to its wide applicability, many variants of PLS have been proposed \citep{Krishnan:2011}. Our approach builds upon a relatively simple form of PLS, which seeks to find the $K$ direction(s) maximizing covariance among two data sets $\bX_1, \bX_2$: 
\begin{equation}
    \argmax_{\bU, \bV} \Tr(\bU\bX_1^{\top}\bX_2\bV) \text{ such that } \bU^{\top}\bU = \bV^{\top}\bV = \bI \label{eqn:pls}
\end{equation}
See the review of \citet{DeBie:2005} for further discussion of this formulation. 

When the features in data sets $\bX_1$ and $\bX_2$ are standardized, this is essentially equivalent to \emph{canonical correlation analysis} (CCA). In fact, the formulation in problem \eqref{eqn:pls} is often referred to as ``diagonal CCA'' and the term PLS is reserved for more complicated formulations. While we focus on the formulation in problem \eqref{eqn:pls}, if the scales of the graph signals are not directly comparable, it may be preferable to use scale-invariant CCA, with the constraint $\bU^{\top}\bX_1^{\top}\bX_1\bU = \bI_k$. 
Various approaches to regularized CCA have been considered in the literature \citep{Witten:2009,Chen:2012,Gao:2017-SCCA,Mohammadi:2017}, though not in the context of graph alignment. 

\subsection{Contributions}
We develop a novel approach to the problem of ``\textbf{Co}arse'' (community-level) \textbf{N}oisy \textbf{G}raph \textbf{A}lignment problem, \textbf{CONGA}: \emph{i.e.}, the problem of identifying related community structures from noisy graph signals on unaligned graphs of potentially different sizes. Unlike vertex assignment based approaches, the method works directly at the group membership level and makes no vertex-to-vertex comparisons. As described in more detail below, the approach flexibly incorporates sparsity and graph-smoothness regularizers for each set of graph signals. Further, it automatically produces paired community estimates as part of its estimation procedure and does not require any downstream processing. 
We provide efficient algorithms to implement the method in Section~\ref{sec:algs}, followed by a discussion of several key theoretical properties in Section~\ref{sec:properties}.  We then demonstrate the effectiveness of the approach through a small simulation study in Section~\ref{sec:sims}, and conclude the paper in Section~\ref{sec:conclusion} with a discussion of several interesting extensions. 

\section{Coarse Graph Alignment via Sparse PLS}
We now introduce our approach to coarse graph alignment from multiple sets of noisy graph signals. As discussed above, our approach combines classical PLS methods with sparsity regularization to capture community structure and graph-smoothness regularization to reflect the observed network structure supporting the graph signals. Taken together, these goals motivate the following optimization problem:
\begin{align}
    \argmax_{\bU, \bV} & \Tr(\bU^{\top}\bX_1^\top\bX_2\bV) - \lambda_1P_1(\bU) - \lambda_2P_2(\bV) \label{eqn:sfpls} \\ 
    \text{subject to }&{\bU \in \textsf{conv}\,\V^{\bI_{n_1} + \alpha_1\bL_1}_{n_1 \times K}, \bV \in \textsf{conv}\,\V^{\bI_{n_2} + \alpha_2\bL_2}_{n_2 \times K}}, \nonumber
\end{align}
where $\mathcal{V}^{\bOmega}_{p \times q}$ denotes the generalized Stiefel manifold of order $p, q$ with respect to the $\bOmega$ product, that is, the set of matrices $\bX \in \R^{p \times q}$ satisfying $\bX^{\top}\bOmega\bX = \bI_q$, $\textsf{conv}\,\mathcal{V}^{\bOmega}_{p \times q} = \{\bX \in \R^{p \times q}: \lambda_{\max}(\bX^{\top}\bOmega\bX) \leq 1\}$ is its convex hull, and $\bL_1, \bL_2$ are the (normalized) Laplacians of $\mathcal{G}_1, \mathcal{G}_2$ respectively. This approach builds upon the regularized Sparse and Functional PCA (SFPCA) framework of \citet{Allen:2019} and its multi-rank extension \citep{Weylandt:2019c}. We highlight several of its compelling features here: 
\begin{itemize}
    \item The left PLS vectors $\bU$ are restricted to the constraint set $\bU^{\top}(\bI + \alpha_1 \bL_1)\bU = \bI$. This imposes a smoothness constraint with respect to the graph Laplacian with strength controlled by the tuning parameter $\alpha_1 \in \R_{\geq 0}$. This type of smoothness is common in GSP and was explored in the graph signal PCA setting by \citet{Jiang:2013}.
    \item We can allow an arbitrary non-smooth penalty $P_1(\cdot)$. In practice, this is typically an $\ell_1$-type penalty to impose node-wise sparsity, \emph{i.e.}, each community corresponds to a subset of nodes, but other penalties can be used as the application demands. 
    \item Heuristically, the combination of orthogonality and sparsity in the PLS vectors induces non-overlapping supports of the columns of $\bU$. This implies that most nodes are only assigned to a single community. 
\end{itemize}
By symmetry, the same properties hold for the right PLS vectors $\bV$; see the paper of \citet{Allen:2019} for more detailed discussion of these properties in the PCA setting. Additionally, the PLS vectors are paired by construction highlighting the corresponding communities of $\mathcal{G}_1$ and $\mathcal{G}_2$.  Finally, note that the two graphs are not required to be of the same size or to have any shared structure beyond the $K$ corresponding communities; the only strong sampling assumption made by our method is $m$ paired signals or, more weakly, the ability to consistently estimate cross-network covariances for each pair of nodes. Figure~\ref{fig:schematic} gives an overview of our model and the interpretation of our \emph{sparse~+~graph-smooth~PLS}~(SGPLS) estimator. 

\subsection{Algorithms} \label{sec:algs}
Next, we discuss of how to implement the SGPLS estimator. The Sparse Graph-Smooth PLS problem \eqref{eqn:sfpls} lacks a closed-form solution when $\lambda_1$ or $\lambda_2$ are non-zero and instead requires the use of an iterative algorithm. We leverage an alternating maximization scheme for problem \eqref{eqn:sfpls}, maximizing over $\bU$ while holding $\bV$ constant and \emph{vice versa}. With $\bV$ held constant, the generalized Stiefel manifold constraints on $\bU$ require the use of non-smooth manifold optimization techniques for each subproblem. Our general approach is outlined in Algorithm~\ref{alg:man_sfpls} below. To solve the $\bU$- and $\bV$-subproblems, we use the Manifold ADMM scheme originally proposed by \citet{Kovantsky:2016}, which alternates between a generalized unbalanced Procrustes problem and a soft-thresholding operator. Given the similarities between problem \eqref{eqn:sfpls} and the multi-rank SFPCA proposal of \citet{Weylandt:2019c}, we refer the reader to that paper for a detailed discussion of the computational details. Note that the manifold proximal gradient schemes of \citet{Chen:2018,Chen:2019} have stronger convergence guarantees and may be used in lieu of the manifold ADMM, but are more expensive to implement and typically converge to inferior stationary points \citep{Weylandt:2019c}.

\begin{algorithm}[ht]
\caption{Sparse Graph-Smooth PLS Algorithm} \label{alg:man_sfpls}
\begin{enumerate}
\item Construct Graph Laplacians $\bL_1, \bL_2$ and associated smoothing matrices $\bS_{1} = \bI_{n_1} + \alpha_{1} \bL_{1}$, $\bS_{2} = \bI_{n_2} + \alpha_{2} \bL_{2}$
  \item Initialize $\hat{\bU}, \hat{\bV}$ to the leading $K$ singular vectors of $\bX_1^{\top}\bX_2$
  \item Repeat until convergence:
  \begin{align*}
      \hat{\bU} &= \argmin_{\bU \in \textsf{conv}\,\V^{\bS_1}_{n_1 \times K}} -\Tr(\bU^{\top}\bX_1^{\top}\bX_2\hat{\bV}) + \lambda_{1}P_{1}(\bU) \\
      \hat{\bV} &= \argmin_{\bV \in \textsf{conv}\,\V^{\bS_2}_{n_2 \times K}} -\Tr(\bV^{\top}\bX_2^{\top}\bX_1\hat{\bU}) + \lambda_{2}P_{2}(\bV)
  \end{align*}
  \item Return $\hat{\bU}$ and $\hat{\bV}$
\end{enumerate}
\end{algorithm}
\noindent For large graphs, the subproblems of Algorithm~\ref{alg:man_sfpls} may be too expensive to directly solve. In this case, we instead recommend the use of Algorithm~\ref{alg:sfpls} which combines a greedy search for each ``block pair'' with a regularized variant of the power method for each subproblem. While this approach does not have the full orthogonality guarantees of Algorithm~\ref{alg:man_sfpls}, it does enforce a weaker form of orthogonality, namely that $\hat{\bu}_k^{\top}\bC_{k + s} = \bzero$ and $\bC_{k+s}\hat{\bv}_k = \bzero$ for $s \geq 0$ where $\bC_1, \bC_2, \dots$ are the ``deflated'' inner-product matrices containing the residual signal after each step \citep[\emph{cf.}][Section 3]{Weylandt:2019c}. This implies that, at each step, all signal attributable to the $k^{\text{th}}$ estimated block is removed and does not reoccur at any subsequent iteration. This does not enforce non-overlap of the estimated block memberships, but, in conjunction with an appropriately tuned sparsity penalty, performs well experimentally.

\begin{algorithm}[ht] 
\caption{Sparse Graph-Smooth PLS Algorithm (Greedy Variant)}\label{alg:sfpls}
\begin{enumerate}
  \item Construct Graph Laplacians $\bL_1, \bL_2$ and associated smoothing matrices $\bS_{1} = \bI_{n_1} + \alpha_{1} \bL_{1}$, $\bS_{2} = \bI_{n_2} + \alpha_{2} \bL_{2}$ with leading eigenvalues $\ell_1 = \lambda_{\max}(\bS_{1})$ and $\ell_2 = \lambda_{\max}(\bS_{2})$
  \item Initialize $\bC_1 \coloneqq \bX_1^{\top}\bX_2$
  \item For $k = 1, \dots K$:
  \begin{enumerate}[label = (\alph*)]
  \item Initialize $\hat{\bu}_k, \hat{\bv}_k$ to the leading singular vectors of $\bC_k$
  \item Repeat until convergence:
  \begin{enumerate}[label = (\roman*)]
    \item $\bu$-subproblem: repeat until convergence: 
    \begin{align*}
    \bu_k &= \prox_{\frac{\lambda_{1}}{\ell_1} P_{1}(\cdot)}\left(\bu_k + \ell_{1}^{-1}\left(\bC_k\hat{\bv}_k - \bS_{1}\bu_k\right)\right) \\
    \hat{\bu}_k &= \begin{cases} \bu_k & \|\bu_k\|_{\bS_{1}} \leq 1 \\ \bu_k / \|\bu_k\|_{\bS_{1}} & \text{ otherwise} \end{cases}
    \end{align*}
    \item $\bv$-subproblem: repeat until convergence: 
    \begin{align*}
    \bv_k &= \prox_{\frac{\lambda_{2}}{\ell_{2}} P_{2}(\cdot)}\left(\bv_k + \ell_{2}^{-1}\left(\bC^{\top}_k\hat{\bu}_k - \bS_{2}\bv_k\right)\right) \\
    \hat{\bv}_k &= \begin{cases} \bv_k & \|\bv_k\|_{\bS_{2}} \leq 1 \\ \bv_k / \|\bv_k\|_{\bS_{2}} & \text{ otherwise} \end{cases}
    \end{align*}
  \end{enumerate}
  \item Set $\bC_{k+1} \coloneqq \bC_k - \frac{\bC_k\hat{\bv}_k\hat{\bu}_k^T\bC_k}{\hat{\bu}_k^T\bC_k\hat{\bv}_k}$
  \end{enumerate}
  \item Return $\{\hat{\bu}_k\}_{k=1}^K$ and $\{\hat{\bv}_k\}_{k=1}^K$
\end{enumerate}
\end{algorithm}

\begin{figure*}[t]
\centering
\includegraphics[width=\textwidth]{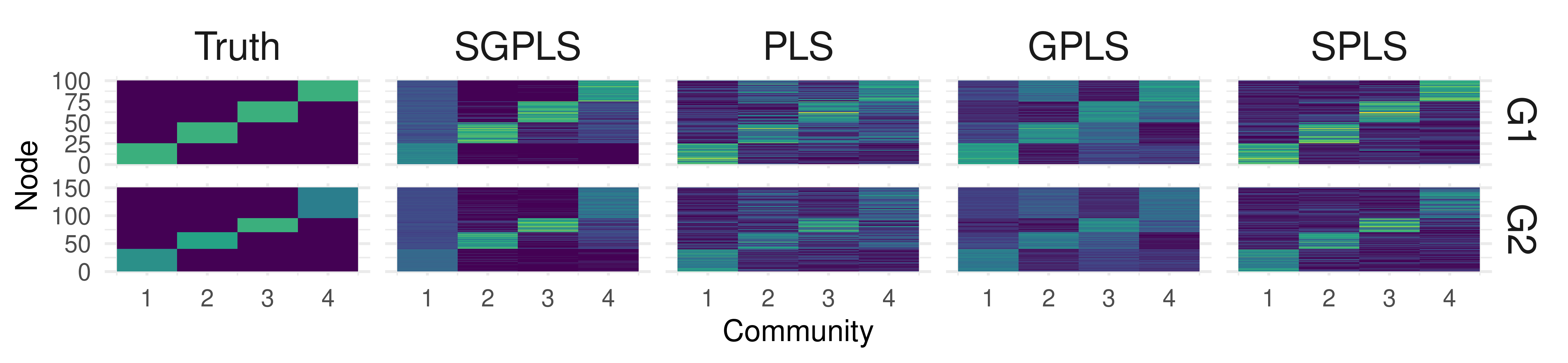}
\vspace{-0.3in}
\caption{Comparison of Sparse Graph-Smooth Partial Least Squares (SGPLS) with unregularized PLS, non-sparse graph-smooth PLS (GPLS), and sparse PLS (SPLS) on the two-way community block detection task described in Section \ref{sec:sims}. SGPLS (second column) is able to most accurately recover the membership matrices for both $\mathcal{G}_1$ (top) and $\mathcal{G}_2$ (bottom). Traditional PLS (third column) does not accurately recover any meaningful group structure. While non-sparse GPLS (fourth column) is able to qualitatively identify some community blocks, the lack of sparsity precludes meaningful separation without agressive down-stream preprocessing. SPLS (fifth column) performs well, but ignores the observed graph structure, and yields a less sparse solution and is consequently more prone to ``false positive'' potential assignments; SGPLS has an advantage here because the graph-Laplacian smoothing ``pools'' several near-zero estimates and pushes them all simultaneously to zero, yielding clearer and more consistent block identification. Oracle tuning parameters were used for each method.}
\label{fig:sims}
\end{figure*}

\subsection{Theoretical Properties} \label{sec:properties}
We state the following theoretical properties of the Sparse Graph-Smooth PLS problem \eqref{eqn:sfpls} without proof. For each result, proofs follow along similar lines as those for the corresponding SFPCA results \citep{Allen:2019}. 

\begin{restatable}{proposition}{degeneracy} \label{prop:degeneracy}
Suppose $P_1, P_2$ are positive homogeneous of order one and let $(\bU^{*}, \bV^{*})$ be the optimal points of problem \eqref{eqn:sfpls}. Then, the following statements hold:
\begin{enumerate}[label=(\roman*)]
  \item There exist values $\lambda_{1}^{\max}$ and $\lambda_{2}^{\max}$ such that, if $\lambda_{1} \geq \lambda_{1}^{\max}$ or if $\lambda_{2} \geq \lambda_{2}^{\max}$, then the solution to problem \eqref{eqn:sfpls} is trivial in the sense $(\bU^*, \bV^*) = (\bzero, \bzero)$.
  \item If $\lambda_{1} < \lambda_{1}^{\max}$ and $\lambda_{2} < \lambda_{2}^{\max}$, the Sparse Graph-Smooth PLS solution $(\bU^{*}, \bV^{*})$ depends smoothly on all (non-zero) regularization parameters.  
  \item $\|\bU^{*}_{\cdot k}\|_{\bS_{1}}$ is equal to either $1$ or $0$ for all $k$, with the latter occurring only when $\lambda_{1} \geq \lambda_{1}^{\max}$ or $\lambda_{2} \geq \lambda_{2}^{\max}$. An analogous result holds for $\bV^{*}$.
  \item $(\bU^{*}, \bV^{*})$ do not suffer from scale non-identifiability. (That is, $(c\bU^*, c^{-1}\bV^*)$ is not a solution for any $c \geq 0$ except $c = 1$.)
\end{enumerate}
\end{restatable}
\noindent At a high level, these properties guarantee that the SGPLS solutions are well-posed and correspond to smoothly varying functions of both the sparsity and smoothness parameters. When degeneracies occur, in the form of empty estimated communities, they are easy to diagnose and rectify. Additionally, we have strong convergence guarantees for each step of the greedy method described in Algorithm~\ref{alg:sfpls}:
\begin{restatable}{proposition}{convergence} \label{prop:convergence}
Suppose $P_1, P_2$ are positive homogeneous of order one, then Algorithm~\ref{alg:sfpls} has the following properties at each $k$:
\begin{enumerate}[label = (\roman*)]
  \item Step 3(b)(i) converges to a stationary point of  
  \begin{equation*} \argmin_{\bu: \bu^{\top}(\bI + \alpha_1\bL_1)\bu \leq 1} \frac{1}{2}\|\bC_k\hat{\bv}_k - \bu\|_2^2 + \lambda_{1} P_{1}(\bu) + \frac{\alpha_{1}}{2}\bu^{\top}\bL_1\bu \label{eqn:sfpls_rank1}\end{equation*}
  Furthermore, if $P_{1}$ is convex, the convergence is monotone, at an $\mathcal{O}(1/K)$ rate, and to a global solution. Step 3(b)(ii) converges analogously for $\bv$ and $P_{2}$. 
  \item If $P_{1}$ is convex, Step 3(b)(i) yields a global solution to 
  \begin{equation} \argmax_{\substack{\bu: \bu^{\top}(\bI + \alpha_1\bL_1) \leq 1 \\ \bv: \bv^{\top}(\bI + \alpha_2\bL_2)\bv \leq 1}} \bu^{\top}\bX\bv - \lambda_{1} P_{1}(\bu) - \lambda_2 P_2(\bv) \label{eqn:sfpls_rank1_constrained} \end{equation} considering $\bC_k$ and $\hat{\bv}_k$ fixed; if $P_{1}$ is non-convex, Step 3(b)(i) yields a stationary point of the same problem, considering $\bC_k $ and $\hat{\bv}_k$ fixed. An analogous result holds for $\hat{\bv}$ returned by Step 3(b)(ii), with $\bC_k$ and $\hat{\bu}_k$ considered fixed.
  \item If $P_{1}, P_{2}$ are both convex, then $(\hat{\bu}_k, \hat{\bv}_k)$ returned by each step 3(b) of Algorithm \ref{alg:sfpls} is both a coordinate-wise global maximum (Nash point) and a stationary point of problem \eqref{eqn:sfpls_rank1_constrained}.
\end{enumerate}
\end{restatable}

\section{Illustration of CONGA on Synthetic Data}
\label{sec:sims}
We demonstrate the efficacy of SGPLS in a small simulation study. We consider two stochastic block models with $K = 4$ communities of sizes $25, 25, 25, 25$ in $\mathcal{G}_1$ ($n_1 = 100$) and sizes $40, 30, 25, 55$ in $\mathcal{G}_2$ ($n_2 = 150$). Each graph was generated with relatively dense intra-community connectivity ($p = 0.95$) and moderately sparse inter-community connectivity ($q = 0.2$). $m = 1000$ paired graph signals were generated by selecting nodes from corresponding communities with probability $s = 0.8$ and assigning a signal with total energy 2 among the selected nodes. Finally, \text{IID} Gaussian noise ($\sigma = 1$) was added at each node, for an effective signal to noise ratio of 0.2 in $\mathcal{G}_1$ and 0.163 in $\mathcal{G}_2$ in each graph signal. 

Figure~\ref{fig:sims} visualizes the results of SGPLS in this situation and compares it to non-regularized, sparse-only, and graph-smooth-only PLS variants. Specifically, we visualize the $\hat{\bU}, \hat{\bV}$ matrices which estimate the cluster membership matrices for each graph. As illustrated in Figure~\ref{fig:sims}, the cluster membership matrices estimated by SGPLS have low degrees of overlap, due to the sparsity and orthogonality constraints, unlike GPLS, and accurately reflect the observed graph structures, unlike SPLS. All PLS variants align communities between the two graphs reasonably well, but SGPLS yields the most accurate coarse graph alignment, in large part because it most accurately identifies communities in the first place.

\section{Discussion and Extensions} \label{sec:conclusion}
We have proposed a sparse partial least squares approach to the problem of coarse graph alignment from noisy signals (CONGA). Our approach builds upon the SVD formulation of the classical PLS problem to incorporate nodewise-sparsity and graph smoothness in order to identify related sets of nodes in disjoint and non-aligned graphs. The combination of two-way sparsity and smoothness allows us to flexibly adapt to different levels of sparsity and smoothness while maintaining non-trivial solutions (\emph{cf.} Proposition \ref{prop:degeneracy}). We have proposed a manifold optimization algorithm to (approximately) identify the optimal pairs of node sets and a greedy approximation suitable for large data sets. Finally, we have demonstrated the statistical and computational efficiency of our approach in simulations. 

At its heart, our method relies on a regularized singular value decomposition of the matrix $\bX_1^{\top}\bX_2$. This perspective suggests several interesting extensions, two of which we highlight next. The SVD minimizes approximation error in the Frobenius norm and is thus suitable when graph signals are corrupted with IID Gaussian noise. For graph signals observed over time, or with other correlation patterns, such as noise arising from a graph diffusion process, a decomposition based on the generalized matrix decomposition of \citet{Allen:2014} may be more suitable. In many situations of interest, it may be suitable to find related blocks among more than two sets of graph signals. In order to adapt our approach to the many-graph setting (as opposed to two graphs), it is necessary to use a multiple PLS or multiple CCA approach, of which several have been proposed in the literature (see, \emph{e.g.}, the discussion in the paper by \citet{Chen:2019-GraphCCA}). For the CONGA problem, a tensor PLS approach of the type considered by \citet{Zhao:2013-HOPLS} is the most natural, in conjunction with the flexible regularization schemes recently developed by \citet{Geyer:2020}. In such settings, all signals may not be observed simultaneously, motivating additional interesting questions for future study.


\clearpage
\section{References}
{\ninept \printbibliography[heading=none]}
\end{refsection}
\end{document}




